\documentclass[10pt,journal,compsoc]{IEEEtran}
\usepackage{xcolor}
\usepackage{amsfonts, amsmath, amssymb, amsthm}

\theoremstyle{definition}

\usepackage{graphicx}
\usepackage{float}
\usepackage{tabularx}
\usepackage{booktabs}
\usepackage{multicol}
\usepackage{multirow}

\usepackage[colorlinks]{hyperref}
\hypersetup{colorlinks,breaklinks,linkcolor=red,urlcolor=blue,anchorcolor=blue,citecolor=blue}

\usepackage{cleveref}

\usepackage{algorithm}
\usepackage{algorithmic}
\usepackage{subfigure}

\DeclareMathOperator*{\argmin}{arg\,min}

\ifCLASSOPTIONcompsoc
  \usepackage[nocompress]{cite}
\else
  \usepackage{cite}
\fi
\ifCLASSINFOpdf
\else
\fi
\begin{document}
%

\title{
Correlating Time Series with Interpretable Convolutional Kernels

}
%
%
%
%

\author{
Xinyu~Chen, HanQin~Cai,~\IEEEmembership{Senior~Member,~IEEE,} Fuqiang~Liu, and~Jinhua Zhao
\IEEEcompsocitemizethanks{\IEEEcompsocthanksitem Xinyu Chen and Jinhua Zhao are with the Department of Urban Studies and Planning, Massachusetts Institute of Technology, Cambridge, MA 02139, USA (e-mail: chenxy346@gmail.com; jinhua@mit.edu).
\IEEEcompsocthanksitem HanQin Cai is with the Department of Statistics and Data Science and Department of Computer Science, University of Central Florida, Orlando, FL 32816 USA (e-mail: hqcai@ucf.edu).
\IEEEcompsocthanksitem Fuqiang Liu is with the Department of Civil Engineering, McGill University, Montreal, QC H3A 0C3, Canada (e-mail: fuqiang.liu@mail.mcgill.ca).
\protect\\
}
\thanks{(Corresponding author: Jinhua Zhao)}}

\IEEEtitleabstractindextext{%
\begin{abstract}

This study addresses the problem of convolutional kernel learning in univariate, multivariate, and multidimensional time series data, which is crucial for interpreting temporal patterns in time series and supporting downstream machine learning tasks. First, we propose formulating convolutional kernel learning for univariate time series as a sparse regression problem with a non-negative constraint, leveraging the properties of circular convolution and circulant matrices. Second, to generalize this approach to multivariate and multidimensional time series data, we use tensor computations, reformulating the convolutional kernel learning problem in the form of tensors. This is further converted into a standard sparse regression problem through vectorization and tensor unfolding operations. In the proposed methodology, the optimization problem is addressed using the existing non-negative subspace pursuit method, enabling the convolutional kernel to capture temporal correlations and patterns. To evaluate the proposed model, we apply it to several real-world time series datasets. On the multidimensional rideshare and taxi trip data from New York City and Chicago, the convolutional kernels reveal interpretable local correlations and cyclical patterns, such as weekly seasonality. In the context of multidimensional fluid flow data, both local and nonlocal correlations captured by the convolutional kernels can reinforce tensor factorization, leading to performance improvements in fluid flow reconstruction tasks. Thus, this study lays an insightful foundation for automatically learning convolutional kernels from time series data, with an emphasis on interpretability through sparsity and non-negativity constraints.

\end{abstract}

}

\maketitle

\IEEEdisplaynontitleabstractindextext

%
\IEEEpeerreviewmaketitle

\section{Introduction}

\IEEEPARstart{T}{ime} series data are one of the most important data types encountered in real-world systems, capturing intrinsic temporal correlations and patterns that are essential for understanding and forecasting various phenomena. Accurately modeling these correlations and patterns is fundamental in many domains, such as spatiotemporal prediction and control systems. To achieve this, it is common to formulate time series coefficients using both linear and nonlinear machine learning approaches, in the meantime providing a flexible framework for analyzing and predicting time-dependent behaviors. In statistics, autoregression (AR) models have been extensively applied to time series analysis, offering an efficient approach to modeling temporal dependencies \cite{box2015time, hamilton2020time}. The classical AR framework also leads to vector autoregression (VAR) for multivariate time series, which captures the interdependencies among a sequence of time series \cite{hamilton2020time}. One classical counterpart that takes the form of VAR---dynamic mode decomposition \cite{tu2013dynamic, schmid2011applications, proctor2016dynamic, brunton2022data}---combines the concepts from fluid dynamics and machine learning to characterize complex dynamical systems. This method is effective in applications such as fluid flow analysis, where it decomposes the dynamics into a set of modes that describe the system's behavior over time.

When applying AR to a circulant time series---where the AR order equals the length of the time series---the AR operation can be equivalently viewed as a circular convolution between the time series and its coefficients. The convolution operation is vital for filtering and signal processing tasks \cite{oppenheim1999discrete}, where the convolution theorem relates the convolution in the time domain to multiplication in the frequency domain via the discrete Fourier transform. Recent advancements in machine learning have further expanded the use of convolution operations in sequence modeling \cite{prince2023understanding}. Convolutional kernel methods such as Laplacian convolutional representation \cite{chen2024laplacian} enable characterizing the complex temporal dependencies. To summarize, the aforementioned regression methods, including AR, VAR, and convolution, take linear equations in an unsupervised learning framework.


Another aspect of enhancing model interpretability in machine learning is using sparsity-induced norms. Typically, structured sparsity regularization offers an effective way to select features and improve model interpretability \cite{jenatton2011structured}. The LASSO method \cite{tibshirani1996regression} is particularly useful for identifying key features when only a subset of features (i.e., input variables) is relevant or correlated with the target variables (i.e., output variables), as it lets the coefficients of irrelevant feature be zero. Since sparsity-inducing norms such as the $\ell_0$- and $\ell_1$-norm enforce sparsity patterns, the resulting algorithms are particularly useful for tasks such as sparse signal recovery \cite{foucart2013invitation}, outlier detection \cite{candes2011robust}, variable selection in genetic fine mapping \cite{wang2020simple}, and nonlinear system identification \cite{brunton2016discovering}, to name but a few.

However, manually designed kernels (e.g., kernels referring to the random walk \cite{cai2010graph}) in time series methods often introduce systematic errors due to human cognitive biases. The kernel learning frameworks vary significantly due to the different purposes such as regression and interpretability. To capture the interpretable kernels for characterizing temporal patterns, we incorporate sparse linear regression into time series convolution with sparse kernels. This study aims to connect time series analysis with the learning process of interpretable convolutional kernels, in which the proposed method offers significant benefits, such as reducing biases in time series convolution and uncovering temporal patterns. Overall, the contribution of this study is three-fold:


\begin{itemize}
\item We reformulate the convolutional kernel learning from univariate time series data as a non-negative $\tau$-sparse regression problem, which is then solved using a greedy method derived from classical subspace pursuit (SP) \cite{dai2009subspace} methods. In the algorithmic implementation, the properties of circulant structure and circular convolution are fully utilized to simplify the computations involved in linear transformation and non-negative least squares.
\item We formulate the $\tau$-sparse regression problem not only for univariate time series but also for multivariate and multidimensional time series, fully utilizing tensor computations. The optimization problem is well-suited for learning convolutional kernels from sequences of time series. Leveraging the properties of tensor computations also allows one to convert the optimization problem involving multivariate or multidimensional time series into standard $\tau$-sparse regression problems.
\item We demonstrate the significance of learning convolutional kernels from several real-world time series datasets, including human mobility data and fluid flow data. The kernels learned from these time series are important for interpreting underlying local and nonlocal temporal correlations and patterns. We empirically show the performance gains by using these convolutional kernels in tensor factorization to address fluid flow reconstruction problems.

\end{itemize}

The remainder of this paper is organized as follows. \Cref{related_work} reviews the related literature, while \Cref{preliminaries} introduces the basic mathematical notations. \Cref{methodologies} presents the $\tau$-sparse regression framework and algorithms for learning convolutional kernels from univariate, multivariate, and multidimensional time series. In \Cref{experiments}, we evaluate the proposed methods on several real-world time series datasets. Finally, we conclude this study in \Cref{conclusion}.


\section{Related Work}\label{related_work}

\subsection{Solving Sparse Regression}

\begin{figure}[h!]
    \centering
    \includegraphics[width=0.4\textwidth]{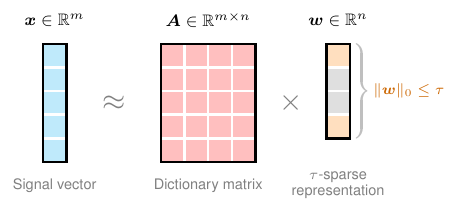}
    \caption{Linear regression problem $\min_{\boldsymbol{w}}~\|\boldsymbol{x}-\boldsymbol{A}\boldsymbol{w}\|_2^2$ with $\tau$-sparse representation of the coefficient vector $\boldsymbol{w}$. In compressive sensing, the goal is to construct a sparse vector $\boldsymbol{w}$ given measurements $\boldsymbol{x}$ (i.e., the signal) and $\boldsymbol{A}$ (e.g., the dictionary) \cite{foucart2013invitation}. The vector $\boldsymbol{w}$ is constrained to have no more than $\tau$ non-zero entries in which $\tau\in\mathbb{Z}^{+}$ refers to the sparsity level.}
    \label{sparse_reg}
\end{figure}

In the fields of signal processing and machine learning, a classical optimization problem involves learning sparse representations \cite{foucart2013invitation} from a linear regression model with measurements $\boldsymbol{x}\in\mathbb{R}^{m}$ and $\boldsymbol{A}\in\mathbb{R}^{m\times n}$, such that
\begin{equation}\label{sparse_reg_eq}
\begin{aligned}
\min_{\boldsymbol{w}}~&\|\boldsymbol{x}-\boldsymbol{A}\boldsymbol{w}\|_2^2 \\
\text{s.t.}~&\|\boldsymbol{w}\|_0\leq\tau,\,\tau\in\mathbb{Z}^{+},
\end{aligned}
\end{equation}
is of great significance for many scientific areas (e.g., compressive sensing \cite{donoho2006compressed, foucart2013invitation}) due to the $\ell_0$-norm on the decision variable $\boldsymbol{w}$, which counts the number of non-zero entries. As shown in \Cref{sparse_reg}, it becomes a classical least squares problem \cite{bishop2006pattern} if $\boldsymbol{x}$ and $\boldsymbol{A}$ are known variables and the vector $\boldsymbol{w}$ is not required to be sparse. In the fields of signal processing and information theory, a large manifold of iterative methods and algorithms have been developed for this problem because the problem \eqref{sparse_reg_eq} is typically NP-hard. These include some of the most classical iterative greedy methods, such as orthogonal matching pursuit (OMP) \cite{pati1993orthogonal, tropp2007signal}, compressive sampling matching pursuit (CoSaMP) \cite{needell2009cosamp}, and subspace pursuit (SP) \cite{dai2009subspace}. Both CoSaMP and SP are fixed-cardinality methods whose support set has a fixed cardinality, while the support set in OMP is appended incrementally during the iterative process. In the case of inferring causality, if the matrix $\boldsymbol{A}$ has $n$ explanatory variables, then the sparse regression problem becomes a classical variable selection technique \cite{wang2020simple}. When $\boldsymbol{w}$ is assumed to be non-negative, the methods derived from OMP, such as non-negative orthogonal greedy algorithms, require one to resolve the non-negative least squares problem \cite{yaghoobi2015fast, nguyen2019non}.

\subsection{Learning Kernels from Time Series}

In the field of statistics, time series problems have been well investigated via the use of AR methods \cite{hamilton2020time}. The coefficients in the AR methods represent the correlations at different times. For certain purposes such as modeling of local temporal dependencies, time series smoothing using random walk can minimize the errors of first-order differencing on the time series. Instead of time series smoothing, Laplacian kernels are more flexible for characterizing the temporal dependencies \cite{chen2024laplacian}, in which the temporal modeling is in the form of a circular convolution between Laplacian kernel and time series. The probability product kernel, constructed based on probabilistic models of the time series data, can evaluate exponential family models such as multinomials and Gaussians and yields interesting nonlinear correlations \cite{jebara2004probability}. The auto-correlation operator kernel can discover the dynamics of time series by evaluating the difference between auto-correlations \cite{chen2013model}. Besides, Gaussian elastic matching kernels possess a time-shift and nonlinear representation in time series analysis~\cite{chen2015kernel}. However, setting the aforementioned kernels requires certain assumptions and prior knowledge, a better way would be to learn the kernels from time series automatically, improving the model interpretability.

\section{Preliminaries}\label{preliminaries}


In this work, we summarize the basic symbols and notations in \Cref{notation}. Here, $\mathbb{R}$ denotes the set of real numbers, while $\mathbb{Z}^{+}$ refers to the set of positive integers. The definitions of tensor unfolding and modal product (or mode-$k$ product as shown in \Cref{notation}) are well explained in \cite{kolda2009tensor, golub2013matrix}. For those symbols and notations related to tensor computations, we also follow the conventions in \cite{kolda2009tensor, golub2013matrix}.

\begin{table}[h!]
\centering
\caption{Summary of the basic notation.}
\label{notation}
\begin{tabular}{l|l} 
\toprule
Notation & Description \\ 
\midrule
$\tau\in\mathbb{Z}^{+}$ & Sparsity level (positive integer) \\
$x\in\mathbb{R}$ & Scalar \\
$\boldsymbol{x}\in\mathbb{R}^{n}$ & Vector of length $n$ \\
$\boldsymbol{X}\in\mathbb{R}^{m\times n}$ & Matrix of size $m\times n$ \\
$\boldsymbol{\mathcal{X}}\in\mathbb{R}^{m\times n\times t}$ & Tensor of size $m\times n\times t$ \\
${\partial f}/{\partial\boldsymbol{X}}$ & Partial derivative of $f$ with respect to $\boldsymbol{X}$ \\
$[i]$ & Positive integer set $\{1,2,\ldots,i\},\,i\in\mathbb{Z}^{+}$ \\
$\|\cdot\|_0$ & $\ell_0$-norm of vector \\
$\|\cdot\|_2$ & $\ell_2$-norm of vector \\
$\|\cdot\|_F$ & Frobenius norm of matrix or tensor \\
$\star$ & Circular convolution \\
$\times_k,\,\forall k\in\mathbb{Z}^{+}$ & Mode-$k$ product between tensor and matrix \\
$\odot$ & Khatri-Rao product \\
\bottomrule
\end{tabular}
\end{table}

In particular, circular convolution is essential when dealing with periodic signals and systems \cite{brunton2022data}, and it is also an important operation in this work. Given two vectors $\boldsymbol{\theta}=(\theta_1,\theta_2,\cdots,\theta_T)^\top\in\mathbb{R}^T$ and $\boldsymbol{x}=(x_1,x_2,\cdots,x_T)^\top\in\mathbb{R}^T$ of length $T$, the circular convolution of $\boldsymbol{\theta}$ and $\boldsymbol{x}$ is denoted by
\begin{equation}
\boldsymbol{y}=\boldsymbol{\theta}\star\boldsymbol{x}\in\mathbb{R}^T,
\end{equation}
element-wise, this gives
\begin{equation}
y_t=\sum_{k\in[T]}\theta_{t-k+1}x_k,\,\forall t\in[T],
\end{equation}
where $y_t$ is the $t$th entry of $\boldsymbol{y}$, and $\theta_{t-k+1}=\theta_{t-k+1+T}$ for $t+1\leq k$. Since the results of circular convolution are computed in a circulant manner, the circular convolution can therefore be rewritten as a linear transformation using a circulant matrix. In this case, we have
\begin{equation}\label{conv_property}
\boldsymbol{y}=\boldsymbol{\theta}\star\boldsymbol{x}=\boldsymbol{x}\star\boldsymbol{\theta}=\mathcal{C}(\boldsymbol{x})\boldsymbol{\theta},
\end{equation}
where $\mathcal{C}:\mathbb{R}^T\to\mathbb{R}^{T\times T}$ denotes the circulant operator \cite{liu2022recovery, chen2024laplacian}. For example, on the vector $\boldsymbol{x}\in\mathbb{R}^T$, the circulant matrix can be written as follows,
\begin{equation}\label{circ_mat_of_x}
\mathcal{C}(\boldsymbol{x})=\begin{bmatrix}
x_1 & x_T & x_{T-1} & \cdots & x_2 \\
x_2 & x_1 & x_{T} & \cdots & x_2 \\
x_3 & x_2 & x_1 & \cdots & x_4 \\
\vdots & \vdots & \vdots & \ddots & \vdots \\
x_T & x_{T-1} & x_{T-2} & \cdots & x_1 \\
\end{bmatrix}\in\mathbb{R}^{T\times T}.
\end{equation}


\section{Methodologies}\label{methodologies}

In this study, we present a convolutional kernel learning method to characterize the temporal patterns of univariate, multivariate, and multidimensional time series data. First, we formulate the optimization problem for learning temporal kernels as a linear regression with sparsity and non-negativity constraints. Then, we solve the optimization problem by using the non-negative SP method.

\subsection{On Univariate Time Series}

\subsubsection{Model Description}

In real-world systems, time series often exhibit complex correlations among both local and nonlocal data points. In this study, we propose characterizing the time series correlations using circular convolution, an approach inspired by the temporal regularization with Laplacian kernels introduced in \cite{chen2024laplacian}. Formally, for the univariate time series $\boldsymbol{x}=(x_1,x_2,\cdots,x_T)^\top\in\mathbb{R}^{T}$ with $T$ time steps, we formulate the learning process as an optimization problem. The objective function involves the circular convolution (denoted by $\star$) between the temporal kernel $\boldsymbol{\theta}$ (i.e., convolutional kernel) and the time series $\boldsymbol{x}$, i.e.,
\begin{equation}\label{sparse_ar_eq}
\begin{aligned}
\min_{\boldsymbol{w}\geq 0}~&\|\boldsymbol{\theta}\star\boldsymbol{x}\|_2^2 \\
\text{s.t.}~&\begin{cases} \boldsymbol{\theta}=\begin{bmatrix} 1 \\ -\boldsymbol{w} \end{bmatrix}, \\ \|\boldsymbol{w}\|_0\leq\tau,\,\tau\in\mathbb{Z}^{+}, 
\end{cases} \\
\end{aligned}
\end{equation}
in which the $(\tau+1)$-sparse kernel $\boldsymbol{\theta}$ is designed to capture temporal correlations. In the parameter setting, we assume the first entry of $\boldsymbol{\theta}$ is $1$, while the remaining $T-1$ entries are non-positive values, parameterized by non-negative vector $\boldsymbol{w}\in\mathbb{R}^{T-1}$. The sparsity constraint applies to $\boldsymbol{w}$, allowing no more than $\tau$ positive entries, where $\tau$ is referred to as the sparsity level. The sparsity assumption is meaningful for parameter pruning, preserving only the most remarkable coefficients to characterize local and nonlocal temporal patterns. In the circular convolution $\boldsymbol{\theta}\star\boldsymbol{x}$ within the objective function, the temporal kernel $\boldsymbol{\theta}$ can also be interpreted as a graph filter with time-shift operator, as seen in the field of graph signal processing (e.g., \cite{ortega2018graph, leus2023graph}). By leveraging the property of circular convolution, $\boldsymbol{\theta}\star\boldsymbol{x}=\boldsymbol{\Theta}\boldsymbol{x}$, the matrix $\boldsymbol{\Theta}\in\mathbb{R}^{T\times T}$ can be expressed as a matrix polynomial:
\begin{equation}
\boldsymbol{\Theta}=\boldsymbol{I}_T-w_1\boldsymbol{F}-w_2\boldsymbol{F}^2-\cdots-w_{T-1}\boldsymbol{F}^{T-1},
\end{equation}
with the $\tau$-sparse representation (i.e., a sequence of coefficients)
\begin{equation}
\boldsymbol{w}=(w_1,w_2,\cdots,w_{T-1})^\top\in\mathbb{R}^{T-1},
\end{equation}
and the time-shift matrix
\begin{equation}
\boldsymbol{F}=\begin{bmatrix}
0 & 0 & 0 & \cdots & 0 & 1 \\
1 & 0 & 0 & \cdots & 0 & 0 \\
0 & 1 & 0 & \cdots & 0 & 0 \\
\vdots & \vdots & \vdots & \ddots & \vdots \\
0 & 0 & 0 & \cdots & 0 & 0 \\
0 & 0 & 0 & \cdots & 1 & 0 \\
\end{bmatrix}\in\mathbb{R}^{T\times T}.
\end{equation}
Herein, $\boldsymbol{I}_T$ is the identity matrix of size $T\times T$. 

As a result, we can express the temporal kernel $\boldsymbol{\theta}$ in the circular convolution $\boldsymbol{\theta}\star\boldsymbol{x}$ and the corresponding circulant matrix $\boldsymbol{\Theta}$ in the matrix-vector multiplication $\boldsymbol{\Theta}\boldsymbol{x}$ as follows,
\begin{equation}\label{theta_w_para}
\boldsymbol{\theta}=(1,-w_1,-w_2,\cdots,-w_{T-1})^\top=\begin{bmatrix}
1 \\ -\boldsymbol{w} \\
\end{bmatrix},
\end{equation}
and 
\begin{equation}
\boldsymbol{\Theta}=\begin{bmatrix}
1 & -w_{T-1} & -w_{T-2} & \cdots & -w_1 \\
-w_1 & 1 & -w_{T-1} & \cdots & -w_2 \\
-w_2 & -w_1 & 1 & \cdots & -w_3 \\
\vdots & \vdots & \vdots & \ddots & \vdots \\
-w_{T-1} & -w_{T-2} & -w_{T-3} & \cdots & 1 \\
\end{bmatrix},
\end{equation}
respectively. Due to the property of circulant matrix, the temporal kernel $\boldsymbol{\theta}$ is indeed the first column of matrix $\boldsymbol{\Theta}$.

\begin{figure}[h!]
    \centering
    \includegraphics[width=0.4\textwidth]{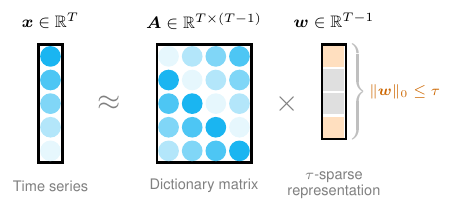}
    \caption{Illustration of learning $\tau$-sparse vector $\boldsymbol{w}$ from the time series data $\boldsymbol{x}$ with the constructed formula as $\boldsymbol{x}\approx\boldsymbol{A}\boldsymbol{w}$. The $T$-by-$(T-1)$ dictionary matrix $\boldsymbol{A}$ is constructed by the time series $\boldsymbol{x}$, see Eq.~\eqref{A_mat}.}
    \label{sparse_reg_time_series}
\end{figure}

As mentioned above, the temporal kernel $\boldsymbol{\theta}$ can be reinforced for capturing local and nonlocal correlations of time series automatically. Using the structure of $\boldsymbol{\theta}$ described in Eq.~\eqref{theta_w_para}, the problem \eqref{sparse_ar_eq} is equivalent to
\begin{equation}\label{optimize_w_ell_0_plus}
\begin{aligned}
\min_{\boldsymbol{w}\geq 0}~&\|\boldsymbol{x}-\boldsymbol{A}\boldsymbol{w}\|_2^2 \\
\text{s.t.}~&\|\boldsymbol{w}\|_0\leq\tau,\,\tau\in\mathbb{Z}^{+},
\end{aligned}
\end{equation}
where the auxiliary matrix $\boldsymbol{A}$ is comprised of the last $T-1$ columns of the circulant matrix $\mathcal{C}(\boldsymbol{x})\in\mathbb{R}^{T\times T}$ (see Eq.~\eqref{circ_mat_of_x}), namely,
\begin{equation}\label{A_mat}
\boldsymbol{A}=\begin{bmatrix}
x_{T} & x_{T-1} & x_{T-2} & \cdots & x_{2} \\
x_{1} & x_{T} & x_{T-1} & \cdots & x_{3} \\
x_{2} & x_{1} & x_{T} & \cdots & x_{4} \\
\vdots & \vdots & \vdots & \ddots & \vdots \\
x_{T-2} & x_{T-3} & x_{T-4} & \cdots & x_{T} \\
x_{T-1} & x_{T-2} & x_{T-3} & \cdots & x_{1} \\
\end{bmatrix}\in\mathbb{R}^{T\times (T-1)}.
\end{equation}
As can be seen, one of the most intriguing properties is the circular convolution $\boldsymbol{\theta}\star\boldsymbol{x}$ can be converted into the expression $\boldsymbol{x}-\boldsymbol{A}\boldsymbol{w}$, which takes the form of linear regression, as illustrated in \Cref{sparse_reg_time_series}. Thus, our problem aligns with sparse linear regression on the data pair $\{\boldsymbol{x},\boldsymbol{A}\}$ in \Cref{sparse_reg}, if not mentioning the non-negativity constraint.

\subsubsection{Solution Algorithm}

To solve the optimization problem in Eq.~\eqref{optimize_w_ell_0_plus}, one should consider both non-negativity and sparsity of the vector $\boldsymbol{w}$. In this study, we present \Cref{sparse_graph_kernel_sp_algo} as the implementation using a non-negative SP method that adapted from \cite{dai2009subspace}, where non-negative least squares is treated as a subroutine. The temporal kernel $\boldsymbol{\theta}$ is constructed using the $\tau$-sparse representation $\boldsymbol{w}$ (see Eq.~\eqref{theta_w_para}). Here, $S=\operatorname{supp}(\boldsymbol{w})=\{t:w_t\neq 0\}$ represents the support set of the vector $\boldsymbol{w}$, with $|S|$ denoting the cardinality of $S$. Notably, we compute $\boldsymbol{a}_i^\top\boldsymbol{r},\,\forall i\in[T-1]$, where the vector $\boldsymbol{a}_i$ is defined as
\begin{equation}\label{vec_ai_univariate}
\boldsymbol{a}_{i}=(x_{T-i+1},\cdots,x_{T},x_1,\cdots,x_{T-i})^\top\in\mathbb{R}^{T},
\end{equation}
where the entries of the first phase start from $x_{T-i+1}$ to $x_T$, as the remaining $T-i$ entries start from $x_1$ to $x_{T-i}$. Such structure is consistent with the matrix $\boldsymbol{A}$ in Eq.~\eqref{A_mat}.

\begin{algorithm}[h]
\caption{Estimating $\boldsymbol{w}$ with non-negative SP}
\label{sparse_graph_kernel_sp_algo}
\begin{algorithmic}[1]
   \STATE {\bfseries Input:} Time series $\boldsymbol{x}\in\mathbb{R}^{T}$, and sparsity level $\tau$ of the sparse representation $\boldsymbol{w}$.
   \STATE Initialize the vector $\boldsymbol{w}:=\boldsymbol{0}$ as zeros, the support set $S:=\emptyset$ as an empty set, and the error $\boldsymbol{r}:=\boldsymbol{x}$.
   \WHILE{not converged}
   \STATE Find $\ell$ as the index set of the $\tau$ largest entries of $|\boldsymbol{A}^\top\boldsymbol{r}|$ in which $\boldsymbol{A}^\top\boldsymbol{r}=(\boldsymbol{a}_1^\top\boldsymbol{r}_1,\boldsymbol{a}_2^\top\boldsymbol{r}_2,\cdots,\boldsymbol{a}_{T-1}^\top\boldsymbol{r}_{T-1})^\top$.
   \STATE Update the support set $S:= S\cup\{\ell\}$.
   \STATE Update the sparse vector $\displaystyle\boldsymbol{w}_{S}:=\argmin_{\boldsymbol{v}\geq0}~\|\boldsymbol{x}-\boldsymbol{A}_{S}\boldsymbol{v}\|_2^2$ with non-negative least squares.
   \STATE Update the support set $S$ as the index set of the $\tau$ largest entries of $|\boldsymbol{w}|$.
   \STATE Set $w_i=0$ for all $i\notin S$.
   \STATE Update the sparse vector $\displaystyle\boldsymbol{w}_{S}:=\argmin_{\boldsymbol{v}\geq0}~\|\boldsymbol{x}-\boldsymbol{A}_{S}\boldsymbol{v}\|_2^2$ with non-negative least squares.
   \STATE Update the error vector $\boldsymbol{r}:=\boldsymbol{x}-\boldsymbol{A}_{S}\boldsymbol{w}_{S}$.
   \ENDWHILE
   \STATE Return the $\tau$-sparse representation $\boldsymbol{w}$.
\end{algorithmic}
\end{algorithm}

In the meantime, let the support set $S=\{\ell_1,\ell_2,\ldots,\ell_{|S|}\}$ represent a sequence of indices, the corresponding sampling matrix $\boldsymbol{A}_{S}\in\mathbb{R}^{T\times |S|}$ is given by
\begin{equation}
\boldsymbol{A}_{S}=\begin{bmatrix}
\mid & \mid & & \mid \\
\boldsymbol{a}_{\ell_{1}} & \boldsymbol{a}_{\ell_2} & \cdots & \boldsymbol{a}_{\ell_{|S|}} \\
\mid & \mid & & \mid \\
\end{bmatrix},
\end{equation}
with the following column vectors:
\begin{equation}
\begin{aligned}
\boldsymbol{a}_{\ell_1}=&~(x_{T-\ell_1+1},\cdots,x_{T},x_1,\cdots,x_{T-\ell_1})^\top, \\
\boldsymbol{a}_{\ell_2}=&~(x_{T-\ell_2+1},\cdots,x_{T},x_1,\cdots,x_{T-\ell_2})^\top, \\
\vdots~& \\
\boldsymbol{a}_{\ell_{|S|}}=&~(x_{T-\ell_{|S|}+1},\cdots,x_{T},x_1,\cdots,x_{T-\ell_{|S|}})^\top. \\
\end{aligned}
\end{equation}
Thus, it suffices to compute the linear transformation
\begin{equation}
\boldsymbol{A}_{S}\boldsymbol{w}_S=\sum_{\ell\in S}w_{\ell}\boldsymbol{a}_{\ell},
\end{equation}
in a memory-efficient manner. Since the matrix $\boldsymbol{A}$ is derived from the circulant matrix $\mathcal{C}(\boldsymbol{x})$, it is possible to avoid explicitly constructing a memory-consuming matrix of size $T\times (T-1)$. In some extreme cases, where the time series is particularly long, directly computing with $T\times (T-1)$ matrices becomes challenging.


\subsection{On Multivariate Time Series}

\subsubsection{Model Description}

For univariate time series, a $\tau$-sparse representation $\boldsymbol{w}\in\mathbb{R}^{T-1}$ can effectively capture temporal correlations and patterns. However, for multivariate time series, the case becomes more complicated because it is unnecessary to learn a separate $\tau$-sparse representation for each individual time series. Instead, a single sparse vector $\boldsymbol{w}\in\mathbb{R}^{T-1}$ is expected to capture consistent correlations and patterns across all time series. For any multivariate time series $\boldsymbol{X}\in\mathbb{R}^{N\times T}$, where $N$ and $T$ are the number and the length of time series, respectively, the learning process of the temporal kernel $\boldsymbol{\theta}$ can be formulated as follows,
\begin{equation}
\begin{aligned}
\min_{\boldsymbol{w}\geq 0}~&\sum_{n\in[N]}\|\boldsymbol{\theta}\star\boldsymbol{x}_{n}\|_2^2 \\
\text{s.t.}~&\begin{cases} \boldsymbol{\theta}=\begin{bmatrix} 1 \\ -\boldsymbol{w} \end{bmatrix}, \\ \|\boldsymbol{w}\|_0\leq\tau,\,\tau\in\mathbb{Z}^{+}, 
\end{cases} \\
\end{aligned}
\end{equation}
where $\boldsymbol{x}_n\in\mathbb{R}^{T}$ is the $n$-th row vector of $\boldsymbol{X}$, corresponding to a single time series. In the objective function, according to the property of circular convolution in Eq.~\eqref{conv_property}, the circular convolution takes the following form:
\begin{equation}
\boldsymbol{\theta}\star\boldsymbol{x}_n=\mathcal{C}(\boldsymbol{x}_n)\boldsymbol{\theta}=\boldsymbol{x}_n-\boldsymbol{A}_n\boldsymbol{w},
\end{equation}
with $\boldsymbol{A}_n\in\mathbb{R}^{T\times (T-1)}$ consisting of the last $T-1$ columns of the circulant matrix $\mathcal{C}(\boldsymbol{x}_n)$. By constructing the matrix $\boldsymbol{A}_n$ for each time series $\boldsymbol{x}_n$ independently, we propose representing $\boldsymbol{A}_n,\,n\in[N]$ as slices of a newly constructed tensor $\boldsymbol{\mathcal{A}}\in\mathbb{R}^{N\times T\times (T-1)}$. Equivalently, we have
\begin{equation}\label{matrix_kernel_learning_optimization}
\begin{aligned}
\min_{\boldsymbol{w}}~&\|\boldsymbol{X}-\boldsymbol{\mathcal{A}}\times_{3}\boldsymbol{w}^\top\|_F^2 \\
\text{s.t.}~&\|\boldsymbol{w}\|_0\leq\tau,\,\tau\in\mathbb{Z}^{+},
\end{aligned}
\end{equation}
where $\times_3$ denotes the modal product along the third mode, namely, mode-3 product. In this case, we have a linear regression with known time series matrix $\boldsymbol{X}$ and dictionary tensor $\boldsymbol{\mathcal{A}}$. The regression expression is particularly written with the modal product.

\begin{figure}[ht!]
    \centering
    \includegraphics[width=0.5\textwidth]{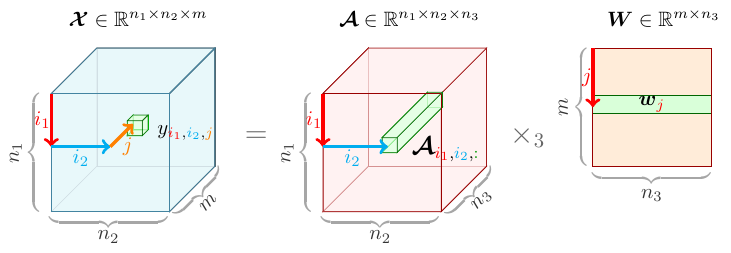}
    \caption{Illustration of the modal product between a third-order tensor and a matrix. By definition, the entry of resulting tensor $\boldsymbol{\mathcal{X}}$ is the inner product between tensor fiber (i.e., in the form of a vector) of $\boldsymbol{\mathcal{A}}$ and row vector of $\boldsymbol{W}$ \cite{kolda2009tensor, golub2013matrix}.}
    \label{mode3_prod}
\end{figure}

\Cref{mode3_prod} illustrates the modal product between any third-order tensor $\boldsymbol{\mathcal{A}}\in\mathbb{R}^{n_1\times n_2\times m}$ and a matrix $\boldsymbol{W}\in\mathbb{R}^{m\times n_3}$. The resulting tensor $\boldsymbol{\mathcal{X}}$ will have dimensions $n_1\times n_2\times m$ by following standard tensor computation principles (see \cite{kolda2009tensor, golub2013matrix} for detailed definitions). If the matrix $\boldsymbol{W}$ is reduced to a row vector, such as the sparse representation $\boldsymbol{w}^\top$ of length $T-1$, then the entries of resulting matrix represent the inner product between the tensor fibers and the vector $\boldsymbol{w}$. In the context of Eq.~\eqref{matrix_kernel_learning_optimization}, the tensor $\boldsymbol{\mathcal{A}}$ is of size $N\times T\times (T-1)$, while the matrix $\boldsymbol{X}$ is of size $N\times T$, allowing for seamless construction of the modal product according to the multiplication principle.

By utilizing the properties of tensor unfolding and modal product as described in \cite{kolda2009tensor}, the optimization problem in Eq.~\eqref{matrix_kernel_learning_optimization} can be equivalently expressed as the following one:
\begin{equation}
\begin{aligned}
\min_{\boldsymbol{w}\geq 0}~&\|\operatorname{vec}(\boldsymbol{X})-\boldsymbol{A}_{(3)}^{\top}\boldsymbol{w}\|_2^2 \\
\text{s.t.}~&\|\boldsymbol{w}\|_0\leq \tau,\,\tau\in\mathbb{Z}^{+},
\end{aligned}
\end{equation}
where $\operatorname{vec}(\cdot)$ denotes the vectorization operator. In tensor computations, $\boldsymbol{A}_{(3)}$ is the tensor unfolding of $\boldsymbol{\mathcal{A}}$ at the third dimension, resulting in $\boldsymbol{A}_{(3)}$ having dimensions $(T-1)\times (NT)$. As can be seen, the original regression problem in Eq.~\eqref{matrix_kernel_learning_optimization} is actually converted into a standard sparse regression problem that is analogous to Eq.~\eqref{optimize_w_ell_0_plus}. Consequently, the previously mentioned algorithm can be applied to the multivariate time series case.

\subsubsection{Solution Algorithm}

Before using \Cref{sparse_graph_kernel_sp_algo}, it is necessary to adjust the algorithm settings. There are some procedures to follow: 1) Set $\boldsymbol{x}:=\operatorname{vec}(\boldsymbol{X})\in\mathbb{R}^{NT}$ as the input. 2) Compute the inner product $\boldsymbol{a}_i^\top\boldsymbol{r},\,\forall i\in[T-1]$, where we have
\begin{equation}\label{vec_ai}
\boldsymbol{a}_i=\bigl(\boldsymbol{x}_{T-i+1}^\top,\cdots,\boldsymbol{x}_{T}^\top,\boldsymbol{x}_{1}^\top,\cdots,\boldsymbol{x}_{T-i}^\top\bigr)^\top\in\mathbb{R}^{NT},
\end{equation}
where $\boldsymbol{x}_{t}\in\mathbb{R}^{N},\,t\in[T]$ are the column vectors of $\boldsymbol{X}\in\mathbb{R}^{N\times T}$. In the vector $\boldsymbol{a}_i$, the entries of the first phase start from $\boldsymbol{x}_{T-i+1}$ to $\boldsymbol{x}_{T}$, as the remaining $N(T-i)$ entries start from $\boldsymbol{x}_{1}$ to $\boldsymbol{x}_{T-i}$. Notably, this principle is analogous to Eq.~\eqref{vec_ai_univariate}.

Finally, suppose $S=\{\ell_1,\ell_2,\ldots,\ell_{|S|}\}$ be the support set, then the most important procedure is constructing the sampling matrix $\boldsymbol{A}_S\in\mathbb{R}^{(NT)\times |S|}$, which consists of the selected columns of $\boldsymbol{A}_{(3)}^\top\in\mathbb{R}^{(NT)\times (T-1)}$ corresponding to the index set $S$. This matrix is given by
\begin{equation}
\boldsymbol{A}_{S}=\begin{bmatrix}
\mid & \mid & & \mid \\
\boldsymbol{a}_{\ell_{1}} & \boldsymbol{a}_{\ell_2} & \cdots & \boldsymbol{a}_{\ell_{|S|}} \\
\mid & \mid & & \mid \\
\end{bmatrix}.
\end{equation}
In this case, if $i\in S$ represents the index $i$ in the support set $S$, then constructing $\boldsymbol{a}_i$ in Eq.~\eqref{vec_ai} allows one to build the column vectors of $\boldsymbol{A}_S$.

\subsection{On Multidimensional Time Series}

For any multidimensional time series $\boldsymbol{\mathcal{X}}\in\mathbb{R}^{M\times N\times T}$ in the form of a tensor, we use the tensor fiber $\boldsymbol{\mathcal{X}}_{m,n,:}\in\mathbb{R}^T$ to represent each individual time series of length $T$. The challenge is to learn a temporal kernel $\boldsymbol{\theta}$ from matrix-variate time series. To address this, we propose formulating the circular convolution as $\boldsymbol{\theta}\star\boldsymbol{\mathcal{X}}_{m,n,:}$ over $m\in[M]$ and $n\in[N]$. Consequently, the optimization problem can be written as follows,
\begin{equation}\label{multi_ts_sparsity_opt}
\begin{aligned}
\min_{\boldsymbol{w}\geq 0}~&\sum_{m\in[M]}\sum_{n\in[N]}\|\boldsymbol{\theta}\star\boldsymbol{\mathcal{X}}_{m,n,:}\|_2^2 \\
\text{s.t.}~&\begin{cases} \boldsymbol{\theta}=\begin{bmatrix} 1 \\ -\boldsymbol{w} \end{bmatrix}, \\ \|\boldsymbol{w}\|_0\leq\tau,\,\tau\in\mathbb{Z}^{+}, 
\end{cases} \\
\end{aligned}
\end{equation}
where the temporal kernel $\boldsymbol{\theta}\in\mathbb{R}^{T}$ is defined such that the first entry is $1$ and the remaining $T-1$ entries are $-\boldsymbol{w}$. Therefore, the optimization problem now becomes
\begin{equation}
\begin{aligned}
\min_{\boldsymbol{w}\geq 0}~&\|\boldsymbol{\mathcal{X}}-\boldsymbol{\mathcal{A}}\times_4\boldsymbol{w}^\top\|_F^2 \\
\text{s.t.}~&\|\boldsymbol{w}\|_0\leq \tau,
\end{aligned}
\end{equation}
where $\boldsymbol{\mathcal{A}}\in\mathbb{R}^{M\times N\times T\times (T-1)}$ is a fourth-order tensor constructed from $\boldsymbol{\mathcal{X}}$. Specifically, the circulant matrix is defined for each time series $\boldsymbol{\mathcal{X}}_{m,n,:}$ independently. Thus, the problem takes the form of tensor regression with known variables being tensors. By utilizing the properties of tensor unfolding and modal product, we can find an equivalent optimization as follows,
\begin{equation}
\begin{aligned}
\min_{\boldsymbol{w}\geq 0}~&\|\operatorname{vec}(\boldsymbol{\mathcal{X}})-\boldsymbol{A}_{(4)}^\top\boldsymbol{w}\|_2^2 \\
\text{s.t.}~&\|\boldsymbol{w}\|_0\leq \tau,
\end{aligned}
\end{equation}
where the vectorization on the third-order tensor $\boldsymbol{\mathcal{X}}$ is $\operatorname{vec}(\boldsymbol{\mathcal{X}})=\operatorname{vec}(\boldsymbol{X}_{(1)})$ with the tensor unfolding of $\boldsymbol{\mathcal{X}}$ at the first dimension being $\boldsymbol{X}_{(1)}\in\mathbb{R}^{M\times (NT)}$. The matrix $\boldsymbol{A}_{(4)}$ is the tensor unfolding of $\boldsymbol{\mathcal{A}}$ at the fourth dimension, which is of size $(T-1)\times (MNT)$.

As mentioned above, the optimization problem can be converted into an equivalent sparse regression on data pair $\{\operatorname{vec}(\boldsymbol{X}),\boldsymbol{A}_{(4)}^\top\}$ using vectorization and tensor unfolding operations. In the algorithmic implementation, the vector $\boldsymbol{a}_i\in\mathbb{R}^{MNT}$ used in inner product $\boldsymbol{a}_i^\top\boldsymbol{r},\,\forall i\in[T-1]$ can be defined as follows,
\begin{equation}
\begin{aligned}
\boldsymbol{a}_i=\bigl(&\operatorname{vec}(\boldsymbol{X}_{T-i+1})^\top,\cdots,\operatorname{vec}(\boldsymbol{X}_{T})^\top, \\
&\operatorname{vec}(\boldsymbol{X}_{1})^\top,\cdots,\operatorname{vec}(\boldsymbol{X}_{T-i})^\top\bigr)^\top\in\mathbb{R}^{MNT}, \\
\end{aligned}
\end{equation}
where $\boldsymbol{X}_{t}\in\mathbb{R}^{M\times N},\,t\in[T]$ are the frontal slices of the tensor $\boldsymbol{\mathcal{X}}\in\mathbb{R}^{M\times N\times T}$. In this case, we introduce the vectorization operation to make the vector $\boldsymbol{a}_i$ identical to the $i$th column of the matrix $\boldsymbol{A}_{(4)}^\top\in\mathbb{R}^{(MNT)\times (T-1)}$. The essential idea of constructing $\boldsymbol{a}_i$ can be generalized to the column vectors of $\boldsymbol{A}_S$ by letting $i\in S$ over a sequence of indices in the support set $S$.

\section{Experiments}\label{experiments}

In this section, we evaluate the proposed method for learning convolutional kernels using real-world time series data. In what follows, we consider several multidimensional time series datasets, including the rideshare and taxi trip data collected from New York City (NYC) and Chicago, which capture human mobility in urban areas, as well as fluid flow dataset that shows temporal dynamics. We use these datasets to identify interpretable temporal patterns and support downstream machine learning tasks, such as tensor completion in fluid flow analysis.

\subsection{On Human Mobility Data}

Human mobility in urban areas typically exhibits highly periodic patterns on a daily or weekly basis, with a significant number of trips occurring during morning and afternoon peak hours and relatively fewer trips during off-peak hours. The NYC TLC trip data provides records of rideshare and taxi trips projected onto the 262 pickup/dropoff zones across urban areas.\footnote{\url{https://www.nyc.gov/site/tlc/about/tlc-trip-record-data.page}.} Each trip is recorded with spatial and temporal information, including pickup time, dropoff time, pickup zone, and dropoff zone. For privacy concerns, the detailed trajectories (e.g., latitude and longitude) of rideshare vehicles and taxis are removed. By aggregating these trips on an hourly basis, the trip data can be represented as mobility tensors such as $\boldsymbol{\mathcal{X}}$ of size $M\times N\times T$, in which the number of zones is $M=N=262$. For numerical experiments, we choose the datasets covering the first 8 weeks starting from April 1, 2024. As a result, the number of time steps is $T=8\times 7\times 24=1344$. 

\begin{figure*}[ht!]
    \centering
    \includegraphics[width=0.85\textwidth]{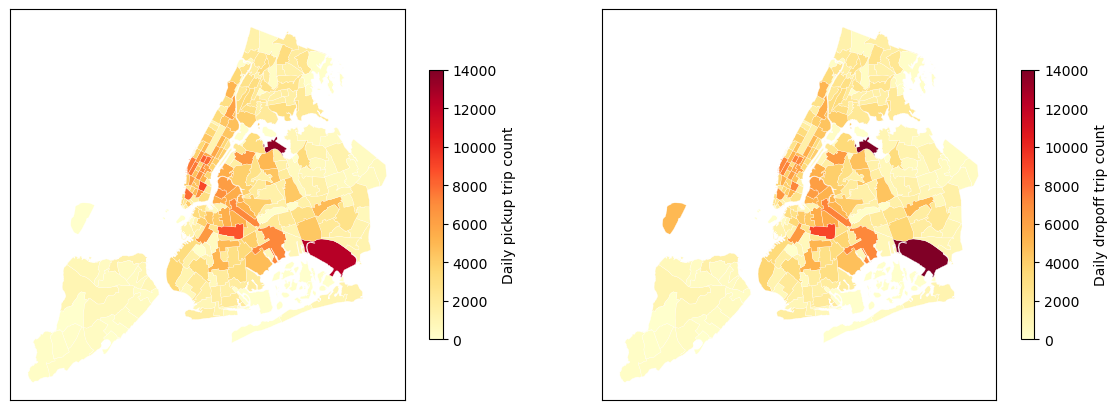}
    \caption{Daily average of rideshare pickup and dropoff trips during the first 8 weeks since April 1, 2024 in NYC, USA. There are 37,404,265 trips in total, while the average daily trips are 667,933.}
    \label{pickup_dropoff_trips_nyc_2024_april_may}
\end{figure*}

\begin{figure}[!ht]
\centering
\subfigure[Total rideshare trips over 262 pickup and dropoff zones.]{
    \centering
    \includegraphics[width = 0.48\textwidth]{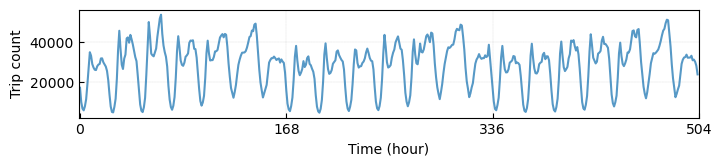}\label{trip_curve_nyc_2024_april_may}
}
\subfigure[Aggregated rideshare pickup trips with the origin as John F. Kennedy International Airport.]{
    \centering
    \includegraphics[width = 0.48\textwidth]{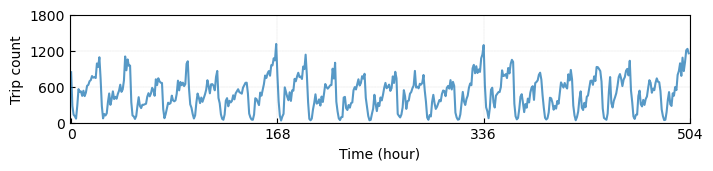}\label{pickup_trip_curve_nyc_132airport_2024_april_may}
}
\subfigure[Aggregated rideshare dropoff trips with the destination as John F. Kennedy International Airport.]{
    \centering
    \includegraphics[width = 0.48\textwidth]{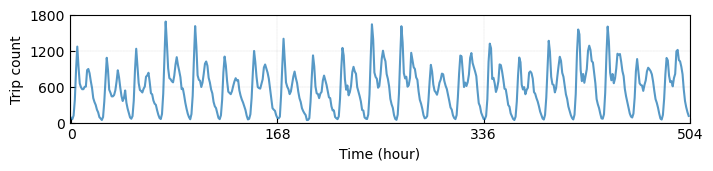}\label{dropoff_trip_curve_nyc_132airport_2024_april_may}
}
\caption{Rideshare trip time series in the first 3 weeks since April 1, 2024 in NYC, USA.}
\end{figure}

\Cref{pickup_dropoff_trips_nyc_2024_april_may} shows the daily average of rideshare pickup and dropoff trips across 262 zones in NYC. From \Cref{trip_curve_nyc_2024_april_may}, one can observe a clear weekly seasonality of rideshare trips in the time series, with similar trends recurring across different weeks. Notably, the airport zones have significantly higher trip counts compared to other zone, as seen in \Cref{pickup_dropoff_trips_nyc_2024_april_may}. As shown in \Cref{pickup_trip_curve_nyc_132airport_2024_april_may,dropoff_trip_curve_nyc_132airport_2024_april_may}, we extract the pickup and dropoff trips associated with John F. Kennedy International Airport. The pickup trip time series shows a distinct trend, peaking every evening, which contrasts with the dropoff trip time series. Nevertheless, both time series exhibit weekly periodic patterns. For comparison, we analyze both rideshare and taxi trip data to highlight the temporal patterns in the experiments.

\begin{figure}[!ht]
\centering
\subfigure[Daily average of pickup and dropoff trips.]{
    \centering
    \includegraphics[width = 0.45\textwidth]{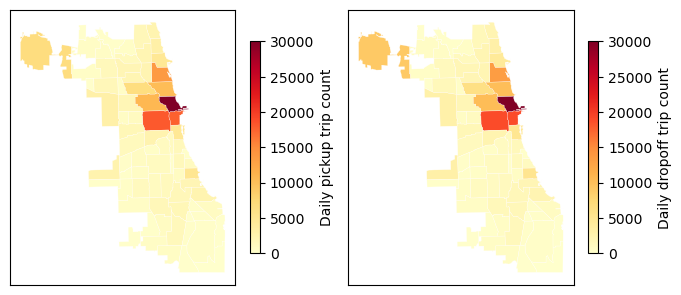}
}
\subfigure[Total rideshare trips over 77 pickup and dropoff zones.]{
    \centering
    \includegraphics[width = 0.48\textwidth]{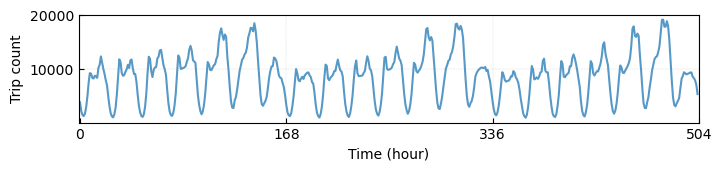}
}
    \caption{Rideshare trips during the first 8 weeks since April 1, 2024 in the City of Chicago, USA. There are 11,374,540 trips in total, while the daily average is 203,117 trips. (a) Pickup and dropoff trips over 77 zones. (b) Aggregated rideshare trips in the first 3 weeks since April 1, 2024.}
\label{pickup_dropoff_trips_chicago_2024_april_may}
\end{figure}

The Chicago Open Data Portal provides the trip records of rideshare vehicles and taxis, mapping onto the 77 pickup/dropoff zones within urban areas.\footnote{\url{https://data.cityofchicago.org/Transportation/Transportation-Network-Providers-Trips-2023-/n26f-ihde/about_data}.}$^,$\footnote{\url{https://data.cityofchicago.org/Transportation/Taxi-Trips-2024-/ajtu-isnz/about_data}.} The trip records can be aggregated into mobility tensors such as $\boldsymbol{\mathcal{X}}$ of size $M\times N\times T$, where $M=N=77$ for the pickup/dropoff zones. We consider both rideshare and taxi data during the first 8 weeks starting from April 1, 2024, comprising $T=1344$ time steps. As shown in \Cref{pickup_dropoff_trips_chicago_2024_april_may}, the time series exhibits clear weekly periodic patterns and consistent time series trends across different weeks.

\begin{table*}[!ht]
\caption{Temporal kernel results achieved by the proposed method on the rideshare and taxi trip datasets in NYC and Chicago. Note that in the first column, ``-R" and ``-T" along with the city refer to the rideshare and taxi, respectively.}
\label{mobility_temporal_kernel_results}
\centering
\begin{tabular}{l|c|c|c}
\toprule
Data & Sparsity & Temporal kernel $\boldsymbol{\theta}\triangleq(1,-\boldsymbol{w}^\top)^\top\in\mathbb{R}^{T}$ & Loss function \\
\midrule
\multirow{4}{*}{NYC-R}
& $\tau=4$ & $(1,\underbrace{\boldsymbol{-0.28}}_{t=2},0,\cdots,0,\underbrace{\boldsymbol{-0.22}}_{t=169},0,\cdots,0,\underbrace{\boldsymbol{-0.22}}_{t=1177},0,\cdots,0,\underbrace{\boldsymbol{-0.28}}_{t=1344})^\top$ & $5.51\times10^7$ \\
& $\tau=6$ & $(1,\underbrace{\boldsymbol{-0.22}}_{t=2},0,\cdots,0,\underbrace{\boldsymbol{-0.14}}_{t=169},0,\cdots,0,\underbrace{\boldsymbol{-0.14}}_{t=337},0,\cdots,0,\underbrace{\boldsymbol{-0.14}}_{t=1009},0,\cdots,0,\underbrace{\boldsymbol{-0.14}}_{t=1177},0,\cdots,0,\underbrace{\boldsymbol{-0.22}}_{t=1344})^\top$ & $5.22\times10^7$ \\
\midrule
\multirow{4}{*}{NYC-T}
& $\tau=4$ & $(1,\underbrace{\boldsymbol{-0.26}}_{t=2},0,\cdots,0,\underbrace{\boldsymbol{-0.23}}_{t=169},0,\cdots,0,\underbrace{\boldsymbol{-0.23}}_{t=1177},0,\cdots,0,\underbrace{\boldsymbol{-0.26}}_{t=1344})^\top$ & $9.69\times10^6$ \\
& $\tau=6$ & $(1,\underbrace{\boldsymbol{-0.20}}_{t=2},0,\cdots,0,\underbrace{\boldsymbol{-0.15}}_{t=169},0,\cdots,0,\underbrace{\boldsymbol{-0.14}}_{t=673},0,\cdots,0,\underbrace{\boldsymbol{-0.14}}_{t=1009},0,\cdots,0,\underbrace{\boldsymbol{-0.15}}_{t=1177},0,\cdots,0,\underbrace{\boldsymbol{-0.20}}_{t=1344})^\top$ & $9.16\times10^6$ \\
\midrule
\multirow{4}{*}{Chicago-R}
& $\tau=4$ & $(1,\underbrace{\boldsymbol{-0.38}}_{t=2},0,\cdots,0,\underbrace{\boldsymbol{-0.13}}_{t=337},0,\cdots,0,\underbrace{\boldsymbol{-0.13}}_{t=1009},0,\cdots,0,\underbrace{\boldsymbol{-0.38}}_{t=1344})^\top$ & $3.23\times10^7$ \\
& $\tau=6$ & $(1,\underbrace{\boldsymbol{-0.36}}_{t=2},0,\cdots,0,\underbrace{\boldsymbol{-0.09}}_{t=337},0,\cdots,0,\underbrace{\boldsymbol{-0.06}}_{t=673},0,\cdots,0,\underbrace{\boldsymbol{-0.09}}_{t=1009},0,\cdots,0,\underbrace{\boldsymbol{-0.06}}_{t=1177},0,\cdots,0,\underbrace{\boldsymbol{-0.36}}_{t=1344})^\top$ & $3.17\times10^7$ \\
\midrule
\multirow{4}{*}{Chicago-T}
& $\tau=4$ & $(1,\underbrace{\boldsymbol{-0.36}}_{t=2},0,\cdots,0,\underbrace{\boldsymbol{-0.15}}_{t=337},0,\cdots,0,\underbrace{\boldsymbol{-0.15}}_{t=1009},0,\cdots,0,\underbrace{\boldsymbol{-0.36}}_{t=1344})^\top$ & $1.74\times10^6$ \\
& $\tau=6$ & $(1,\underbrace{\boldsymbol{-0.30}}_{t=2},0,\cdots,0,\underbrace{\boldsymbol{-0.10}}_{t=25},0,\cdots,0,\underbrace{\boldsymbol{-0.11}}_{t=337},0,\cdots,0,\underbrace{\boldsymbol{-0.11}}_{t=1009},0,\cdots,0,\underbrace{\boldsymbol{-0.10}}_{t=1321},0,\cdots,0,\underbrace{\boldsymbol{-0.30}}_{t=1344})^\top$ & $1.64\times10^6$ \\
\bottomrule
\end{tabular}
\end{table*}

In what follows, we use both NYC and Chicago datasets in the form of tensors to test the proposed method for learning interpretable convolutional kernels. \Cref{mobility_temporal_kernel_results} summarizes the temporal kernels with different sparsity levels $\tau=4$ and $6$ on both rideshare and taxi in the two cities. These temporal kernels reveal the most significant correlations between adjacent time steps, such as $t=1$ and $t=2$ (forward direction) or $t=1344$ (backward direction). The rideshare/taxi trip data of Chicago shows stronger local correlations than the NYC data. When the sparsity level $\tau$ is set to $6$, the temporal kernel captures both nearest time steps $t=2,1344$ and time steps related to weekly seasonality, such as $t=169,337,1009,1177$ in the NYC rideshare dataset and $t=337,673,1009,1177$ in the Chicago rideshare dataset. In addition, using a relatively greater $\tau$ in the convolutional kernel learning process contributes to the reduction of loss functions, in which the loss function corresponds to the objective function in Eq.~\eqref{multi_ts_sparsity_opt}.

Furthermore, it is also meaningful to examine the differences among the weights $\{-\boldsymbol{w}_1,-\boldsymbol{w}_2,\cdots,-\boldsymbol{w}_{T-1}\}$ (i.e., the last $T-1$ entries of $\boldsymbol{\theta}$) of the temporal kernels in \Cref{mobility_temporal_kernel_results}. On the one hand, the temporal kernels for these datasets capture the weekly or bi-weekly seasonality. For instance, the temporal kernel with $\tau=6$ for the NYC rideshare dataset shows consistent weights for weekly and bi-weekly time steps. On the other hand, comparing the temporal kernels across the four different datasets reveals the following findings:
\begin{itemize}
\item \emph{Comparability of temporal kernels}. The local and nonlocal temporal patterns across different datasets are comparable with respect to the weights of temporal kernels. Although these datasets exhibit complicated spatiotemporal correlations, the intrinsic patterns such as weak seasonality can be clearly revealed by the proposed method. For example, the proposed method ($\tau=4$) learns the same support set $S$ in the sparse representation $\boldsymbol{w}$ for the NYC rideshare and taxi data, while the value of weights in $\boldsymbol{w}$ are very close.
\item \emph{Rideshare and taxi trips in NYC exhibit similar strengths of weekly seasonality}, with the sum of nonlocal weights ($\tau=6$) being $-0.56$ for the rideshare dataset against $-0.58$ for the taxi dataset.
\item \emph{Taxi trips in Chicago show stronger weekly seasonality than rideshare trips}, as the sum of nonlocal weights ($\tau=4$) is $-0.30$ for the taxi dataset, compared to $-0.26$ for the rideshare dataset.
\item \emph{Taxi trips in Chicago reveal both daily and weekly seasonality when $\tau=6$}, with the sum of nonlocal weights being $-0.42$ on the taxi dataset, compared to $-0.30$ for the rideshare dataset, indicating stronger seasonality in taxi trips.
\item \emph{NYC trip datasets display stronger seasonality than Chicago trip datasets}. For instance, when $\tau=6$, the sums of nonlocal weights of NYC rideshare, NYC taxi, Chicago rideshare, and Chicago taxi are $-0.56$, $-0.58$, $-0.30$, and $-0.42$, respectively. Similar evidence is seen with $\tau=4$, where the sum of nonlocal weights is $-0.44$ for the NYC rideshare dataset, compared to $-0.26$ for the Chicago rideshare dataset.
\end{itemize}
Therefore, the absolute values of nonlocal weights in the temporal kernels provide a way to measure the periodicity of urban human mobility across different cities and various transportation modes, such as rideshare vehicles and taxis. Since the kernel learning mechanism automatically captures temporal correlations and patterns, these temporal kernels offer valuable insights into real-world systems. Given the consistent settings, such as time periods and transportation modes used in selecting the datasets, the findings discussed above are crucial for policymaking in urban systems.

\begin{table}[!ht]
\caption{The local and nonlocal coefficients in the sparse representation $\boldsymbol{w}\in\mathbb{R}^{T-1}$ on the NYC rideshare datasets from 2019 to 2024. Note that the sparsity level is set as $\tau=4$.}
\label{NYC_kernels}
\centering
\begin{tabular}{l|cccccccc}
\toprule
\multirow{2}{*}{Year} & \multicolumn{8}{c}{Support set $S=\{\ell_1,\ell_2,\ldots,\ell_{|S|}\}$} \\
\cmidrule{2-9}
& 1 & 24 & 168 & 336 & 1008 & 1176 & 1320 & 1343 \\
\midrule
2019 
& 0.27 & 0 & 0.22 & 0 & 0 & 0.22 & 0 & 0.27 \\
2020
& 0 & 0.23 & 0.23 & 0 & 0 & 0.23 & 0.23 & 0 \\
2021
& 0 & 0 & 0.24 & 0.23 & 0.23 & 0.24 & 0 & 0 \\
2022
& 0.26 & 0 & 0.23 & 0 & 0 & 0.23 & 0 & 0.26 \\
2023
& 0.27 & 0 & 0.22 & 0 & 0 & 0.22 & 0 & 0.27 \\
2024
& 0.28 & 0 & 0.22 & 0 & 0 & 0.22 & 0 & 0.28 \\
\bottomrule
\end{tabular}
\end{table}

For complementary needs, \Cref{NYC_kernels} summarizes the $\tau$-sparse representation $\boldsymbol{w}\in\mathbb{R}^{1343}$ achieved by the proposed method on the NYC rideshare data from the first 8 weeks starting April 1st across different years. The results show consistent temporal correlations in 2019, 2022, 2023, and 2024. Specifically, local time steps and weekly seasonality are observed in the support set $S=\{1,168,1176,1343\}$, with the entries of $\boldsymbol{w}$ being remarkably consistent across these years. In 2020, the $\tau$-sparse representation reveals both daily and weekly seasonality in the support set $S=\{24,168,1176,1320\}$, significantly differing from 2019 due to the impact of the COVID-19 pandemic. In 2021, the $\tau$-sparse representation also highlights strong nonlocal patterns such as weekly seasonality in the support set $S=\{168,336,1008,1176\}$. These findings imply that NYC rideshare trips exhibit more periodic patterns during the COVID-19 years.

\subsection{On Fluid Flow Data}

\subsubsection{Learning Convolutional Kernels}

\begin{figure*}[ht!]
    \centering
    \includegraphics[width=0.99\textwidth]{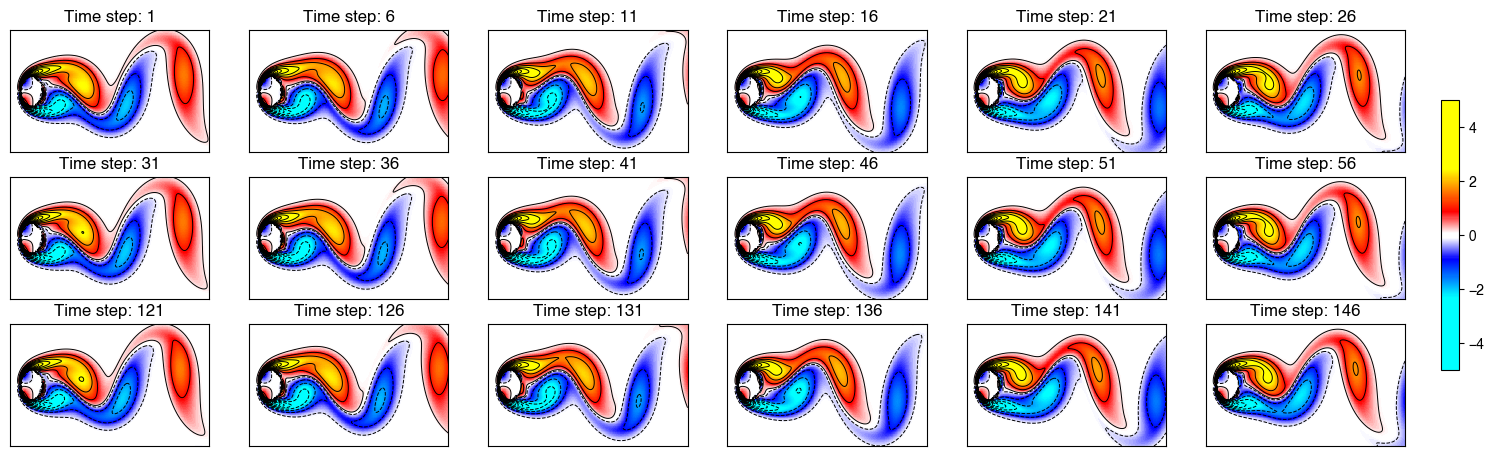}
    \caption{Matrix-variate time snapshots of the fluid flow dataset. This fluid flow dataset has the seasonality $\Delta t=30$. To demonstrate the periodic patterns, the time snapshots since $t=121$ are also presented.}
    \label{fluid_flow_snapshots}
\end{figure*}

The dynamics of fluid flow often exhibit complicated spatiotemporal patterns, allowing one to interpret convolutional kernels in the context of temporal dynamics. We use a fluid flow dataset collected from the fluid flow passing a circular cylinder with laminar vortex shedding at Reynolds number, using direct numerical simulations of the Navier-Stokes equations.\footnote{\url{http://dmdbook.com/}.} This dataset is a multidimensional tensor of size $199\times 449\times 150$, representing $199$-by-$449$ vorticity fields with $150$ time snapshots as shown in \Cref{fluid_flow_snapshots}.

\Cref{fluid_flow_temporal_kernel_results} summarizes the temporal kernels achieved by Algorithm~\ref{sparse_graph_kernel_sp_algo} on the fluid flow dataset with different sparsity levels $\tau=2,3,4$. When $\tau=2$, the temporal kernel $\boldsymbol{\theta}$ primarily captures local correlations between the nearest time snapshots. As the sparsity level increases to $\tau=3,4$, the temporal kernels also capture seasonal patterns at $t=31,121$, reflecting cyclical temporal dynamics in addition to local correlations at $t=2,150$. These temporal kernels enable the correlation of time snapshots in fluid flow data, it is therefore important to examine the significance of convolutional kernels in tensor factorization for addressing the fluid flow reconstruction problem.

\begin{table*}[!ht]
\caption{Temporal kernel results achieved by the proposed method on the fluid flow dataset.}
\label{fluid_flow_temporal_kernel_results}
\centering
\begin{tabular}{l|c|c|c}
\toprule
Sparsity & Temporal kernel $\boldsymbol{\theta}\triangleq(1,-\boldsymbol{w}^\top)^\top\in\mathbb{R}^{T}$ & Loss function & Correlation \\
\midrule
$\tau=2$ & $(1,\underbrace{\boldsymbol{-0.50}}_{t=2},0,\cdots,0,\underbrace{\boldsymbol{-0.50}}_{t=150})^\top$ & $2.49\times10^4$ & Local \\
\midrule
$\tau=3$ & $(1,\underbrace{\boldsymbol{-0.40}}_{t=2},0,\cdots,0,\underbrace{\boldsymbol{-0.21}}_{t=121},0,\cdots,0,\underbrace{\boldsymbol{-0.40}}_{t=150})^\top$ & $2.10\times10^4$ & Local \& nonlocal \\
\midrule
$\tau=4$ & $(1,\underbrace{\boldsymbol{-0.35}}_{t=2},0,\cdots,0,\underbrace{\boldsymbol{-0.16}}_{t=31},0,\cdots,0,\underbrace{\boldsymbol{-0.16}}_{t=121},0,\cdots,0,\underbrace{\boldsymbol{-0.35}}_{t=150})^\top$ & $1.91\times10^4$ & Local \& nonlocal \\
\bottomrule
\end{tabular}
\end{table*}

\subsubsection{Fluid Flow Reconstruction with Tensor Factorization}

For any partially observed tensor $\boldsymbol{\mathcal{Y}}\in\mathbb{R}^{M\times N\times T}$ in the form of multidimensional time series, we consider the problem of fluid flow reconstruction using CP tensor factorization which is a classical formula in tensor computations \cite{kolda2009tensor, golub2013matrix}. To emphasize the significance of learning convolutional kernels from time series data, we reformulate the optimization problem of tensor factorization by incorporating spatiotemporal regularization terms such that
\begin{equation}\label{tf_opt}
\begin{aligned}
\min_{\boldsymbol{W},\boldsymbol{U},\boldsymbol{V}}~&\frac{1}{2}\bigl\|\mathcal{P}_{\Omega}\bigl(\boldsymbol{\mathcal{Y}}_{(1)}-\boldsymbol{W}(\boldsymbol{V}\odot\boldsymbol{U})^\top\bigr)\bigr\|_F^2 \\
&+\frac{\gamma}{2}\bigl(\|\boldsymbol{\Theta}_{w}\boldsymbol{W}\|_F^2+\|\boldsymbol{\Theta}_{u}\boldsymbol{U}\|_F^2+\|\boldsymbol{\Theta}_{v}\boldsymbol{V}\|_F^2\bigr), \\
\end{aligned}
\end{equation}
where $\boldsymbol{\mathcal{Y}}_{(1)}$ is the mode-1 tensor unfolding of size $M\times(NT)$, and $\Omega$ denotes the observed index set of $\boldsymbol{\mathcal{Y}}_{(1)}$. Since the data is partially observed, $\mathcal{P}_{\Omega}(\cdot)$ denotes the orthogonal projection supported on $\Omega$, while $\mathcal{P}_{\Omega}^\perp(\cdot)$ denotes the orthogonal projection supported on the complement of $\Omega$. In this tensor factorization, given a rank $R\in\mathbb{Z}^{+}$, there are three factor matrices $\boldsymbol{W}\in\mathbb{R}^{M\times R}$, $\boldsymbol{U}\in\mathbb{R}^{N\times R}$, and $\boldsymbol{V}\in\mathbb{R}^{T\times R}$. Accordingly, if one accounts for the temporal correlations, the matrix $\boldsymbol{\Theta}_v\in\mathbb{R}^{T\times T}$ is the circulant matrix with the first column being the temporal kernel $\boldsymbol{\theta}_v\in\mathbb{R}^{T}$. Instead of temporal kernel $\boldsymbol{\theta}_v$, the proposed method can also learn the spatial kernels $\boldsymbol{\theta}_w$ and $\boldsymbol{\theta}_u$ from the fluid flow data. Thus, one can construct the spatial regularization terms with matrices $\boldsymbol{\Theta}_w\in\mathbb{R}^{M\times M}$ and $\boldsymbol{\Theta}_u\in\mathbb{R}^{N\times N}$. Notably, these regularization terms are weighted by $\gamma\in\mathbb{R}$.

The optimization problem in Eq.~\eqref{tf_opt} can be solved by the alternating minimization method, in which the variables $\{\boldsymbol{W},\boldsymbol{U},\boldsymbol{V}\}$ would be updated iteratively with the following principle:
\begin{equation}
\begin{cases}
\boldsymbol{W}:=\{\boldsymbol{W}\mid\partial f/\partial\boldsymbol{W}=\boldsymbol{0}\}, \\
\boldsymbol{U}:=\{\boldsymbol{U}\mid\partial f/\partial\boldsymbol{U}=\boldsymbol{0}\}, \\
\boldsymbol{V}:=\{\boldsymbol{V}\mid\partial f/\partial\boldsymbol{V}=\boldsymbol{0}\},
\end{cases}
\end{equation}
where the objective function is denoted by $f$. Each subproblem can be resolved by the conjugate gradient method efficiently \cite{chi2019nonconvex}.

In \Cref{fluid_reconstruction_results}, we randomly generate missing entries with certain missing rates as 50\%, 70\%, and 90\% in the fluid flow $\boldsymbol{\mathcal{X}}$ and construct a partially observed tensor $\boldsymbol{\mathcal{Y}}$ as the input for tensor factorization. We denote the estimated tensor by $\hat{\boldsymbol{\mathcal{Y}}}$ and use the relative squared error as $\text{RSE}=\|\mathcal{P}_{\Omega}(\hat{\boldsymbol{\mathcal{Y}}}-\boldsymbol{\mathcal{X}})\|_2/\|\mathcal{P}_{\Omega}(\boldsymbol{\mathcal{X}})\|_2\times 100$ to measure the imputation performance. To highlight the importance of convolutional kernels, we consider the rank as $R=100$ in different settings of tensor factorization:
\begin{itemize}
\item \textbf{(TF).} Tensor factorization with $\gamma=0$, implying no regularization term.
\item \textbf{(TF-$\boldsymbol{\theta}_v$).} Tensor factorization with the convolutional kernel $\boldsymbol{\theta}_{v}\in\mathbb{R}^{T}$ in which the sparsity level is $\tau=4$ as shown in \Cref{fluid_flow_temporal_kernel_results}. Herein, the weight is set as $\gamma=1\times10^3$.
\item \textbf{(TF-$\{\boldsymbol{\theta}_w,\boldsymbol{\theta}_v\}$).} Tensor factorization with the convolutional kernels $\boldsymbol{\theta}_w\in\mathbb{R}^{M}$ of sparsity level $\tau=2$ and $\boldsymbol{\theta}_v\in\mathbb{R}^{T}$ of sparsity level $\tau=4$. Here, the weight is set as $\gamma=1\times10^1$.
\item \textbf{(TF-$\{\boldsymbol{\theta}_u,\boldsymbol{\theta}_v\}$).} Tensor factorization with the convolutional kernels $\boldsymbol{\theta}_u\in\mathbb{R}^{N}$ of sparsity level $\tau=2$ and $\boldsymbol{\theta}_v\in\mathbb{R}^{T}$ of sparsity level $\tau=4$. Here, the weight is set as $\gamma=1\times10^1$.
\item \textbf{(TF-$\{\boldsymbol{\theta}_w,\boldsymbol{\theta}_u,\boldsymbol{\theta}_v\}$).} Tensor factorization with convolutional kernels $\{\boldsymbol{\theta}_w,\boldsymbol{\theta}_u,\boldsymbol{\theta}_v\}$ in which the spatial kernels $\boldsymbol{\theta}_w\in\mathbb{R}^{M}$ and $\boldsymbol{\theta}_u\in\mathbb{R}^{N}$ are with sparsity level $\tau=2$ and the temporal kernel $\boldsymbol{\theta}_v\in\mathbb{R}^{T}$ is with sparsity level $\tau=4$. Here, the weight is set as $\gamma=1\times10^{-4}$.
\end{itemize}
Of the results in \Cref{fluid_reconstruction_results}, the tensor factorization with temporal kernel $\boldsymbol{\theta}_v$ performs better than the purely tensor factorization, highlighting the importance of temporal kernels. As we have multiple kernel settings in tensor factorization, the performance of fluid flow reconstruction can be further improved when introducing spatial kernels such as $\boldsymbol{\theta}_w$ and $\boldsymbol{\theta}_u$ along the spatial dimensions of fluid flow data.

\begin{table}[!ht]
\caption{Performance (RSE) of the fluid flow reconstruction with tensor factorization methods. The missing values with a certain missing rate are generated 20 times with different random seeds, while the results are given in average and standard deviation of RSEs.}
\label{fluid_reconstruction_results}
\centering
\begin{tabular}{l|cccc}
\toprule
\multirow{2}{*}{Model} & \multicolumn{3}{c}{Missing rate} \\
\cmidrule{2-4}
& 50\% & 70\% & 90\% \\
\midrule
TF 
& $2.30\pm0.10$ & $2.43\pm0.14$ & $3.40\pm0.23$ \\
TF-$\boldsymbol{\theta}_v$
& $2.26\pm0.10$ & $2.40\pm0.13$ & $3.29\pm0.17$ \\
TF-$\{\boldsymbol{\theta}_w,\boldsymbol{\theta}_v\}$
& $2.27\pm0.10$ & $\boldsymbol{2.21}\pm0.11$ & $\boldsymbol{2.64}\pm0.13$ \\
TF-$\{\boldsymbol{\theta}_u,\boldsymbol{\theta}_v\}$
& $2.30\pm0.10$ & $2.42\pm0.14$ & $3.24\pm0.20$ \\
TF-$\{\boldsymbol{\theta}_w,\boldsymbol{\theta}_u,\boldsymbol{\theta}_v\}$
& $\boldsymbol{2.22}\pm0.12$ & $2.24\pm0.11$ & $\boldsymbol{2.64}\pm0.21$ \\
\bottomrule
\end{tabular}
\end{table}

\section{Conclusion}\label{conclusion}

In this study, we propose a unified machine learning framework for temporal convolutional kernel learning to model univariate, multivariate, and multidimensional time series data and capture interpretable temporal patterns. Specifically, the optimization problem for learning temporal kernels is formulated as a linear regression with $\tau$-sparsity (i.e., using $\ell_0$-norm on the sparse representation $\boldsymbol{w}$) and non-negativity constraints. The temporal kernel $\boldsymbol{\theta}$ takes the first entry as one and the remaining entries as $-\boldsymbol{w}$. To ensure the interpretable temporal kernels, the constraints in optimization are solved by the non-negative SP method, which is well-suited to produce a sparse and non-negative sparse representation $\boldsymbol{w}$.

In the modeling process, the challenge arises as the time series switched from univariate cases to multivariate and even multidimensional cases due to the purpose of learning a single kernel $\boldsymbol{\theta}$ from a sequence of time series. To address this, we propose formulating the optimization problem with tensor computations, involving both modal product and tensor unfolding operations in tensor computations. Eventually, we show that the optimization for multivariate and multidimensional time series can be converted into an equivalent sparse regression problem. Thus, the non-negative SP method can be seamlessly adapted for solving these complex optimization problems.

Through evaluating the proposed method on the real-world human mobility data, we show the interpretable temporal kernels for characterizing multidimensional rideshare and taxi trips in both NYC and Chicago, allowing one to uncover the local and nonlocal temporal patterns such as weekly periodic seasonality. The comparison between different cities and transportation modes provides insightful evidence for understanding the periodicity of urban systems. On the fluid flow data, convolutional kernels that obtained along spatial and temporal dimensions can reinforce the tensor completion in fluid flow reconstruction problems.

Although this work focuses on how to learn a temporal kernel from univariate, multivariate, and multidimensional time series data, the essential idea can be easily generalized to other machine learning tasks on relational data. For future work, possible directions for extending the proposed methods include: 1) Learning convolutional kernels from sparse or irregular time series due to the challenge of biased sampling of data points. 2) Inferring causality from time series data.



\appendices

\ifCLASSOPTIONcompsoc
  \section*{Acknowledgments}
\else
  \section*{Acknowledgment}
\fi

This research is based upon work supported by the U.S. Department of Energy’s Office of Energy Efficiency and Renewable Energy (EERE) under the Vehicle Technology Program Award Number DE-EE0009211 and DE-EE0011186. The views expressed herein do not necessarily represent the views of the U.S. Department of Energy or the United States Government. The Mens, Manus, and Machina (M3S) is an interdisciplinary research group (IRG) of the Singapore MIT Alliance for Research and Technology (SMART) center. The work of H.Q.~Cai is partially supported by NSF DMS 2304489. 


\ifCLASSOPTIONcaptionsoff
  \newpage
\fi



%

\bibliographystyle{IEEEtran}
\bibliography{references}

%


\begin{IEEEbiography}[{\includegraphics[width=1in,height=1.25in,clip,keepaspectratio]{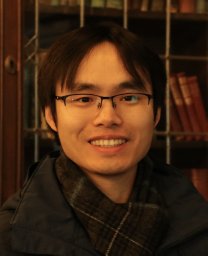}}]{Xinyu Chen} received his Ph.D. degree from the University of Montreal, Montreal, QC, Canada. He is now a Postdoctoral Associate at Massachusetts Institute of Technology, Cambridge, MA, United States. His current research centers on machine learning, spatiotemporal data modeling, intelligent transportation systems, and urban science. \end{IEEEbiography}

\begin{IEEEbiography}[{\includegraphics[width=1in,height=1.25in,clip,keepaspectratio]{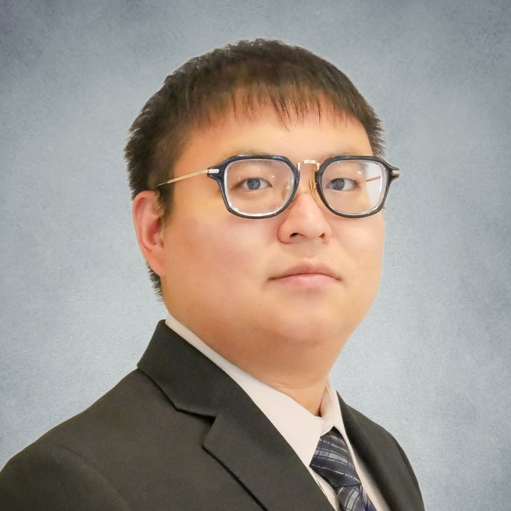}}]{HanQin Cai} (Senior Member, IEEE) received the PhD degree in applied mathematics and computational sciences from the University of Iowa. He is currently the Paul N. Somerville Endowed assistant professor with the Department of Statistics and Data Science and the Department of Computer Science, University of Central Florida. He is also the director of Data Science Lab. His research interests include machine learning, data science, mathematical optimization, and applied harmonic analysis.
\end{IEEEbiography}

\begin{IEEEbiography}[{\includegraphics[width=1in,height=1.25in,clip,keepaspectratio]{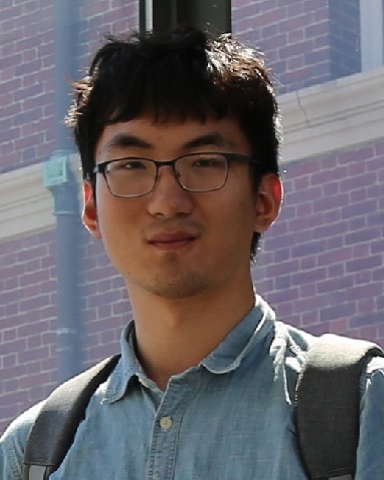}}]{Fuqiang Liu} (Student Member, IEEE) is currently pursuing the Ph.D. degree with the Department of Civil Engineering, McGill University, Montreal,
Canada. His research interests include spatiotemporal data analysis, adversarial studies of deep learning, the robustness of intelligent transportation systems, and the efficient design of deep neural networks.
\end{IEEEbiography}

\begin{IEEEbiography}[{\includegraphics[width=1in,height=1.25in,clip,keepaspectratio]{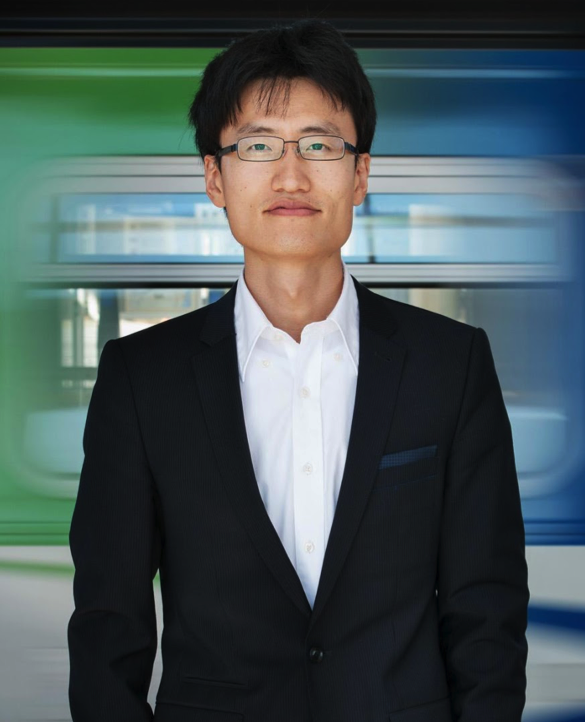}}]{Jinhua Zhao}
is currently the Edward H. and Joyce Linde Associate Professor of city and transportation planning at MIT. He brings behavioral science and transportation technology together to shape travel behavior, design mobility systems, and reform urban policies. He directs the MIT Urban Mobility Laboratory and Public Transit Laboratory.
\end{IEEEbiography}


\vfill


\end{document}